\documentclass[conference,a4paper]{IEEEtran}

\usepackage[utf8]{inputenc} 
\usepackage[T1]{fontenc}
\usepackage{url}
\usepackage{ifthen}
\usepackage{cite}
\usepackage[cmex10]{amsmath}
\usepackage[linesnumbered,ruled]{algorithm2e}
\usepackage{graphicx}
\usepackage{amsmath,amssymb}
\usepackage{subcaption}

\DeclareMathOperator*{\argmin}{arg\,min}
\DeclareMathOperator*{\supp}{supp}
\DeclareMathOperator*{\ber}{Ber}
\DeclareMathOperator*{\SNR}{SNR}

\newcommand{\RR}{\mathbb{R}}

\newcommand{\EE}{\mathbb{E}}
\newcommand{\PP}{\mathbb{P}}
\newcommand{\vecx}{{\bf x}}
\newcommand{\vecw}{{\bf w}}
\newcommand{\vecv}{\mathbf{v}}
\newcommand{\vecy}{\mathbf{y}}
\newcommand{\veca}{\mathbf{a}}

\newcommand{\vecf}{\mathbf{f}}
\newcommand{\vecg}{\mathbf{g}}

\newcommand{\matG}{\mathbf{G}}
\newcommand{\matA}{\mathbf{A}}

\newcommand{\matP}{\mathbf{P}}
\newcommand{\matQ}{\mathbf{Q}}

\newcommand{\ones}{{\bf 1}}
\newcommand{\zero}{{\bf 0}}

\newcommand{\ie}{{\it i.e.,} }
\newcommand{\err}{\operatorname{err}}

\newcommand{\mcA}{\mathcal{A}}

\newtheorem{theorem}{Theorem}[section]
\newtheorem{cor}[theorem]{Corollary}

\newtheorem{lemma}[theorem]{Lemma}
\newtheorem{definition}[theorem]{Definition}

\begin{document}
\title{Gradient Coding via the Stochastic Block Model} 

\author{%
\IEEEauthorblockN{Zachary Charles}
\IEEEauthorblockA{University of Wisconsin-Madison\\
              Department of Electrical and Computer Engineering\\
              Madison, WI 53706\\
              Email: zcharles@wisc.edu}
\and
\IEEEauthorblockN{Dimitris Papailiopoulos}
\IEEEauthorblockA{University of Wisconsin-Madison\\
              Department of Electrical and Computer Engineering\\
              Madison, WI 53706\\
              Email: dimitris@ece.wisc.edu}
}

\maketitle

\begin{abstract}
Gradient descent and its many variants, including mini-batch stochastic gradient descent, form the algorithmic foundation of modern large-scale machine learning. Due to the size and scale of modern data, gradient computations are often distributed across multiple compute nodes. Unfortunately, such distributed implementations can face significant delays caused by straggler nodes, \ie nodes that are much slower than average. Gradient coding is a new technique for mitigating the effect of stragglers via algorithmic redundancy. While effective, previously proposed gradient codes can be computationally expensive to construct, inaccurate, or susceptible to adversarial stragglers. In this work, we present the stochastic block code (SBC), a gradient code based on the stochastic block model. We show that SBCs are efficient, accurate, and that under certain settings, adversarial straggler selection becomes as hard as detecting a community structure in the multiple community, block stochastic graph model.
\end{abstract}

\section{Introduction}

In order to scale up machine learning  to large data sets, we often use distributed implementations of traditional first-order optimization algorithms. While distributed algorithms can theoretically achieve substantial speedups, in practice the parallelization gains often fall short of the optimal speedup gains predicted by theory\cite{dean2012large,paleo}.
One of the sources of this phenomenon is the presence of {\it stragglers}. These are compute nodes whose runtime is much higher than the runtime of most other nodes in the system.

There has been significant recent work in reducing the effect of stragglers in distributed algorithms. Current approaches include replicating jobs across nodes\cite{shah2016redundant} and dropping stragglers in settings where the system can tolerate errors \cite{ananthanarayanan2013effective}. More recently, coding theory has gained traction as a way to speed up distributed. It has been used to reduce the runtime of the shuffling phase of MapReduce \cite{li2015coded}, make distributed matrix multiplication more efficient \cite{dutta2016short}, and reduce the effect of stragglers in certain machine learning computations \cite{lee2016speeding}.

Gradient coding, a straggler-mitigating technique for distributed gradient-based methods, was first proposed in \cite{tandon2016gradient} and was later extended in \cite{raviv2017gradient}. While gradient coding was initially used for exact reconstruction of the sum of gradients, it was later extended to approximate reconstructions using expander and Ramanujan graphs \cite{raviv2017gradient}. Such graphs can be expensive to compute in practice, especially for large numbers of compute nodes, and the approximation error that they introduce during the gradient recovery problem can be large. 
 
Some efficient and simple approximate gradient codes were studied in \cite{charles2017approximate}, including fractional repetition codes (FRCs) and Bernoulli gradient codes (BGCs). FRCs were first presented in \cite{tandon2016gradient}, but only for exact gradient coding. FRCs achieve very small approximation error when the straggler are chosen at random, but suffer when the stragglers are selected adversarially. However, \cite{charles2017approximate} shows that adversarial straggler selection is NP-hard in general, which implies the existence of gradient codes that might be robust to adversarial stragglers under polynomial-time bounded adversaries assuming some widely accepted computational conjectures.

The main problem we tackle in this paper is whether it is possible to design gradient codes that 1) are efficiently computable, 2) achieve reconstruction error similar to FRCs for random stragglers, 3) can be resistant to adversarial stragglers.

	\subsection{Our Contributions}

		In this work, we present the {\it stochastic block code} (SBC), a new gradient code that combines the small reconstruction error of FRCs under random straggler selection with the randomness of BGCs, which is some cases leads to robustness against adversarial stragglers. SBCs are based on the stochastic block model from random graph theory. The code is more effective than BGCs when stragglers are chosen randomly, but more difficult for an adversary to attack than FRCs. We give explicit bounds on the approximation error of SBCs under random straggler selection and show that certain adversarial attacks on SBCs are computationally as hard as recovery problems in community detection for regimes where no polynomial time algorithm is known to exist. Finally, we empirically evaluate SBCs and show that they are capable of achieving small approximation error, even when a constant fraction of the compute nodes are stragglers.

\section{Preliminaries}

	In this work we use standard script for scalars and bold for vectors and matrices. Given a matrix $\matA$, let $\matA_{i,j}$ denote its $(i,j)$ entry and $\veca_j$ denote its $j$th column. Let $\ones_m$ denote the $m\times 1$ all ones vector, while let $\ones_{n \times m}$ denote the all ones $n\times m$ matrix. We define $\zero_m$ and $\zero_{n\times m}$ analogously. Given a matrix $\matA$, let $\matA^+$ denote its pseudoinverse.

\begin{figure}
	\centering
	\includegraphics[width=0.9\columnwidth]{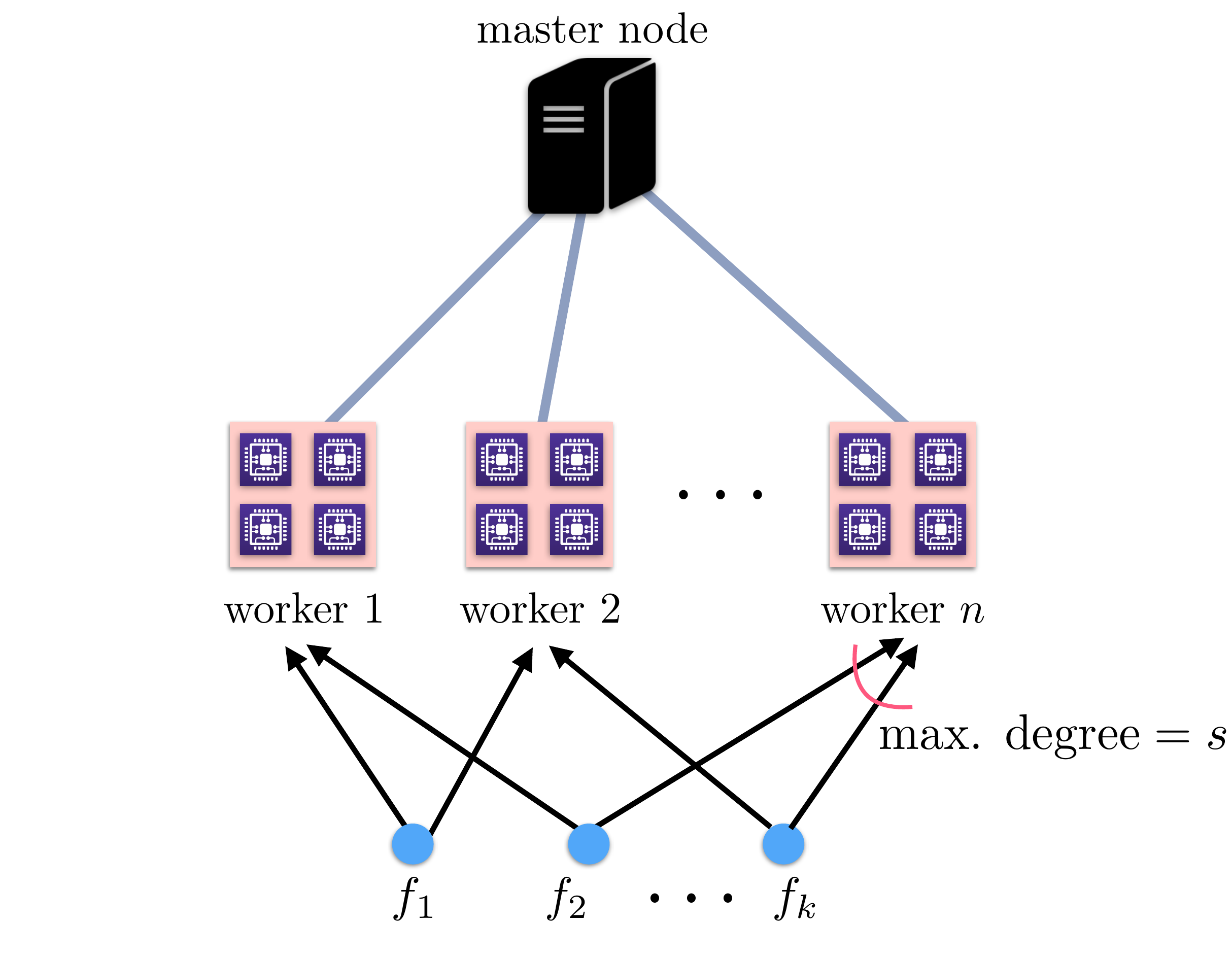}
	\caption{A master-worker distributed system where each compute node has multiple cores. }\label{fig:system}
	\end{figure}

	We consider a distributed master-worker setup of $n$ compute nodes, each of which is assigned $s$ tasks. A pictorial description is given in Figure \ref{fig:system}.
	Each compute node locally computes a set of assigned functions and sends a linear combination of these functions to the master node. 

	The goal of the master node is to compute the sum of $k$ functions
	\begin{equation}\label{eq:sum_functions}
	f({\bf x}) = \sum_{i=1}^k f_i({\bf x})	 = {\bf f}^T{\bf 1_k}
	\end{equation}
	in a distributed way, where $f_i:\RR^d\to\RR^w$ and each $f_i$ can be assigned to and computed locally by any of the $n$ compute nodes. Here, ${\bf f} := [f_1(\vecx),\ldots, f_k(\vecx)]^T$. We denote the output of compute node $j$ by $y_j$.

	Due to the straggler effect, we assume that the master only has access to the output of $r < n$ non-straggler compute nodes. By not waiting for the output of straggler nodes, we can drastically improve the runtime of distributed algorithms. If we wish to exactly recover $f(\vecx)$, then we need $r \geq k-s+1$ \cite{tandon2016gradient}. However, in practice we may only need to approximately recover $f(\vecx)$, which we can do with significantly fewer non-straggler nodes.

	The above setup is relevant to distributed learning algorithms, where we often wish to find some model $\vecx$ by minimizing
	$$\ell(\vecx) = \sum_{i=1}^k \ell(\vecx;{\bf z}_i).$$
	Here $\{{\bf z}_i\}_{i=1}^k$ are our training samples and $\ell(\vecx;{\bf z})$ is a loss function measuring the accuracy of the model $\vecx$ with respect to the data point ${\bf z}$. 
	In order to find a model that minimizes the sum of losses $\ell(\vecx)$, we often use first-order, or gradient-based, methods. In these algorithms, at every distributed iteration, we would need to compute the gradient of $\ell(\vecx)$ (or a mini-batch of the training samples), given  by
	$\nabla \ell(\vecx) = \sum_{i=1}^k \nabla \ell(\vecx;{\bf z}_i).$
	Letting $f_i(\vecx) = \nabla \ell(\vecx;{\bf z}_i)$, we arrive at the setup in (\ref{eq:sum_functions}). Note that this kind of setup applies to both mini-batch SGD and full-batch gradient descent.

	\subsection{Gradient Codes}

		A {\it gradient code} involves a choice of function assignments per compute node, messages sent from the compute node to the master, and a {\it decoding} method used by the master node to approximately reconstruct the true gradient. For simplicity, we assume that the master node can only compute linear combinations of the output of the non-straggler compute nodes.

		Suppose we have $k$ functions to compute and $k$ compute nodes. We wish to construct a $k \times k$ {\it function assignment matrix} $\matG$ that describes which functions each node computes and what message the node passes back to the master node. The column $\vecg_j$ corresponds to compute node $j$. The entries of column $j$ correspond to the coefficients of the linear combination of functions that the compute node sends back to the master node.
		In other words, if column $j$ has support $U$, then node $j$ computes $f_i$ for $i \in U$ and has output $y_j = \sum_{i \in U} f_i$ (or more generally, a linear combination of the computed functions). Let $\vecy = [y_1,\ldots, y_k]^T$.

		For the decoding, we assume that we have $r$ non-straggler nodes with indices given by $T$. Decoding the gradient code corresponds to designing some vector $\vecv \in \RR^k$ with support in $T$. The approximation $\tilde{f}$ to $f$ is then given by
		$$\tilde{f} = \sum_{i=1}^k y_i\vecv_i = \vecy^T\vecv.$$
		We define the error of our approximation in terms of the vector $\vecv$. Note that we have
		\begin{align*}
		(\tilde{f}-f)^2 &= \|\vecf^T\matG\vecv - \vecf^T\ones_k\|_2^2 \leq \|\vecf\|_2^2\|\matG\vecv-\ones_k\|_2^2.\end{align*}
		Since the error depends on $\vecf$ which we do not know a priori, we instead define the approximation error $\err(\vecv)$ of our gradient code by
		$$\err(\vecv) := \|\matG\vecv-\ones_k\|_2^2$$
		where $\vecv$ has support only among the non-straggler nodes.

		Let $T$ denote the indices of the non-straggler columns. We define the $k \times k$ non-straggler matrix $\matG_{T}$ as having the same entries as $\matG$ in the non-straggler columns and having 0 in the straggler columns. This implies
		$\err(\vecv) = \|\matG_{T}\vecv-\ones_k\|_2^2.$
		The optimal decoding vector $\vecv_{\text{opt}}$ is defined by
		$$\vecv_{\text{opt}} := \argmin_{\vecv}\|\matG_{T}\vecv-\ones_k\|_2^2 = \argmin_{\vecv : \supp(\vecv) \subseteq T}\|\matG\vecv-\ones_k\|_2^2.$$
		Standard facts about the pseudoinverse of a matrix then imply
 		$\vecv_{\text{opt}} = \matG_{T}^+\ones_k.$
 		This is one of the possibly infinite many optimal solutions to the above least squares recovery problem.
		In practice, it may not be feasible to compute $\vecv_{\text{opt}}$ since it involves computing the pseudo-inverse of a matrix. Moreover, $\vecv_{\text{opt}}$ is difficult to analyze theoretically since it involves the pseudoinverse of a random matrix. For these two reasons, we will often use and analyze simpler, more efficient decoding methods.

\section{Gradient Coding Methods}

	In this section, we briefly discuss two previously proposed gradient codes and their properties.

	\subsection{Fractional Repetition Codes}

		Fractional repetition codes (FRCs) were first proposed in \cite{tandon2016gradient} for exact reconstruction of $f$. They were later shown in \cite{charles2017approximate} to be useful in the context of approximate gradient coding. Suppose we are given $k$ tasks and $k$ workers and a desired number $s$ of tasks per worker, where we assume $s | k$ without loss of generality. We define the $k\times k$ function assignment matrix $\matG$ of an FRC by
		$${\bf G} = \begin{pmatrix} \ones_{s\times s} & {\bf 0}_{s\times s} & \ldots & {\bf 0}_{s\times s}\\
		{\bf 0}_{s\times s} & \ones_{s\times s} & \ldots & {\bf 0}_{s\times s}\\
		\vdots & \vdots & \ddots & \vdots\\
		{\bf 0}_{s\times s} & {\bf 0}_{s\times s} & \ldots & \ones_{s\times s}\end{pmatrix}.$$
		Note that if columns 1 and 2 are both non-stragglers, then using both columns does not help our decoding. To decode, we first look at indices $1,\ldots, s$. If any are non-stragglers, we select the first such index $t$ and let $\vecv_t = 1$. We repeat this with indices $s+1,\ldots, 2s$, etc. As long as there is a non-straggler in each of these blocks, a simple calculation shows that $\err(\vecv) = 0$.

		FRCs have small error with high probability, if the stragglers are selected uniformly at random, even with a constant fraction of stragglers \cite{charles2017approximate}. However, the worst case error of FRCs is quite large, as we discuss in Section \ref{sec:adv}. To remedy this, we use gradient codes that employ randomness.

	\subsection{Bernoulli Gradient Codes}

		Bernoulli gradient codes (BGCs) were first introduced in \cite{charles2017approximate}. We construct $\matG$ where $\matG_{i,j}$ is Bernoulli with probability $p$. We typically take $p = s/k$, so that each node computes roughly $s$ functions. While we could use optimal decoding and find $\vecv_{\text{opt}}$, there are more efficient (though slightly less accurate) decoding methods. If $T$ is the set of non-stragglers where $|T| = r$, then somewhat small approximation error can be achieved by setting $\vecv_i = \frac{k}{rs}$ if $i \in T$, and 0 otherwise. In \cite{charles2017approximate}, it was shown that under this decoding, with high probability the error satisfies $\err(\vecv) = O(k/rp)$. While less accurate than FRCs, BGCs do not seem to be as vulnerable to adversarial straggler selection.

\section{Stochastic Block Codes}

	Our goal is to construct a gradient code that combines the small error to random stragglers of FRCs and the potential resilience against polynomial-time adversaries of BGCs. We propose a kind of interpolation between the two codes that we refer to as the {\it stochastic block code} (SBC). For such codes, we use a function assignment matrix ${\bf G}$ with block structure
	\begin{equation}\label{sbc}\matG = \begin{pmatrix} \matP & \matQ & \ldots & \matQ\\
	\matQ & \matP &\ldots  & \matQ\\
	\vdots & \vdots & \ddots & \vdots \\
	\matQ & \matQ &  \ldots & \matP\end{pmatrix}.\end{equation}

	Here, each $\matP, \matQ$ is an $s\times s$ matrix with Bernoulli random entries. The entries of each $\matP$ are Bernoulli with probability $p$, while the entries of each $\matQ$ are Bernoulli with probability $q$. In other words, $\matG$ has a stochastic block model structure with $\frac{k}{s}$ partitions of size $s$ with probabilities $p$ within a community and probabilities $q$ between communities. In the following, we will assume $p \geq q$.

	Typically we will consider the case where $p$ is close to $1$ and $q$ is small, so that $\matG$ has blocks similar to $\ones_{s\times s}$ on the diagonal and only a small number of ones outside of the diagonal blocks. This model specializes FRCs when $q = 1, p = 0$ and BGCs when $q = p$. By varying $p$ and $q$, we can interpolate between FRCs and BGCs.

	\subsection{Decoding}

		We would like to devise a computationally efficient method for decoding. Let $S_1 = \{1,\ldots, s\}, S_2 = \{s+1,\ldots, 2s\},\ldots S_{k/s} = \{k-s+1,\ldots, k\}$. Given the set of non-straggler indices $T$, let $T_i = S_i \cap T$. Our decoding method works as follows. For each $T_i \neq \emptyset$, we will select $s_i \in T_i$ at random. We will then add up the columns ${\bf g}_{s_i}$ to form a vector $\vecw$ and then scale $\vecw$ so that it is close to $\ones_k$. The idea is that if we select one column from each block, then their sum should be close to a multiple of the all ones vector. A formal algorithm is given in Algorithm \ref{alg:sbc_decode}.

		\begin{algorithm}{\small
			\SetKwInOut{Input}{Input}
			\SetKwInOut{Output}{Output}
			\Input{The indices $T_1,\ldots, T_{k/s}$ of non-straggler nodes.}
			\Output{A vector $\vecv$ that describes the decoding strategy.}
			$\vecv = \zero_k$\;
			\uIf{$p+(\frac{k}{s}-1)q < 2$}{
				$\beta = 1$\;
			}
			\Else{
				$\beta = p+(\frac{k}{s}-1)q$\;
			}
			\For{$i = 1$ \KwTo $k/s$}{
				\If{$T_i \neq \emptyset$}{
					Pick $t_i \in T_i$ uniformly at random\;
					$\vecv_{t_i} = 1/\beta$\;
				}
			}
			return $\vecv$\;
			\caption{Stochastic block decoding.}
			\label{alg:sbc_decode}
		}
		\end{algorithm}

		Recall that if we want to use $\vecv$ to compute an approximation to $f$, we take $\tilde{f} = \vecy^T\vecv$, where $\vecy$ is the vector of outputs of the nodes. Since $\vecv$ only has non-zero indices corresponding to non-straggler nodes, we can compute $\tilde{f}$ only using the non-straggler nodes.

		This decoding corresponds to selecting one non-straggler column (if it exists) from each block of $s$ columns and then adding these columns together. We select at most $\frac{k}{s}$ columns. Let $\vecw$ denote the sum of the selected columns. We return $\beta^{-1}\vecw$, where $\beta$ is the expected value of each entry of $\vecv$. When $p = 1, q = 0$, this method gives us the optimal decoding method for FRCs discussed above.

		When $p \approx 1$ and $q < \frac{s}{k}$, scaling by $\beta^{-1}$ is not necessarily beneficial. In this regime, each entry of $\vecw$ is 1 with high probability, while a few entries may be larger integers if any of the $\ber(q)$ entries are non-zero. Scaling by $\beta^{-1}$ then introduces errors in to all the entries that are $1$ in $\vecw$. Intuitively, if $X$ is 1 with high probability and 2 with small probability, the variable $Y = X/\EE[X]$ has expected value 1, but will never equal 1, while $X$ will equal 1 with high probability. Thus, not scaling by $\beta$ implies $\vecv \approx \ones_k$ with high probability and has the added benefit of simplifying the analysis.

		One could improve this algorithm by averaging over the non-straggler columns in each $T_i$, and then adding up the averages. When $r$ is large, this approach is better than stochastic block decoding, at the expense of being more difficult to analyze theoretically.

	\subsection{Random Stragglers}\label{sec:random_stragglers}

		Suppose we use an SBC with stochastic block decoding and that the $r$ non-straggler indices $T$ are selected uniformly at random from all subsets of $\{1,\ldots, k\}$ of size $r$. We are particularly interested in the setting where the number of stragglers is a constant fraction of $k$, as this is when previous theory such as that in \cite{tandon2016gradient} breaks down.

		Because of the straggler effect, we only have access to the matrix $\matG_T$ where zeros have been filled in to all straggler columns. Note that $\matG$ is a random matrix, but even fixing $\matG$, $\matG_T$ is random. While the worst-case error may still be large, as it is for FRCs, the average case is much better. We have the following theorem about the reconstruction error.

		\begin{theorem}\label{thm:main_thm}Suppose that $s \geq 2\ln(k)k/r$, $1-p \leq k^{-2}$ and that $q = \frac{\gamma}{k}$ for $\gamma < s$. If we apply stochastic block decoding to a SBC with randomly selected stragglers, then with probability at least $1-\frac{4}{k}$, this produces a vector $\vecv$ such that
		$$\err(\vecv) \leq 14\sqrt{\dfrac{\ln(k)\gamma}{s}}\max\left\{k\left((1-p)+\frac{\gamma}{s}\right),3\ln(k)\right\}.$$
		\end{theorem}
	All our proofs can be found in the long version of this paper \cite{charles2018gradient}.

		Letting $p = 1$, we get the following corollary concerning SBCs that are close to FRCs.
		\begin{cor}\label{cor:frc}Suppose that $s \geq 2\ln(k)k/r$ and $q = \frac{\gamma}{k}$ for $\gamma \leq \frac{3\ln(k)s}{k}$. If the stragglers are selected randomly, then with probability at least $1-\frac{3}{k}$, applying stochastic block decoding method to such a SBC produces a vector $\vecv$ such that
		$$\err(\vecv) \leq O\left(\frac{\ln^2(k)}{\sqrt{k}}\right).$$\end{cor}

		The above theorem required $q < \frac{s}{k}$. When $(\frac{k}{s}-1)q \gg p$, we derive the following theorem.
		\begin{theorem}\label{thm:main_thm2}Suppose that $s \geq 2\ln(k)k/r$ and that \\$(\frac{k}{s}-1)(1-q)q \geq \ln(k)$. If the stragglers are selected randomly, then with probability at least $1-\frac{2}{k}$, applying stochastic block decoding to a SBC produces a vector $\vecv$ satisfying
		$$\err(\vecv) \leq \frac{16\ln(k)s^2}{kq^2}\left((1-p)p + \left(\dfrac{k}{s}-1\right)(1-q)q\right).$$\end{theorem}	

		By setting $p = q$, we can analyze the error of BGCs under stochastic block decoding.
		\begin{cor}\label{cor:bgc}Suppose that $s \geq 2\ln(k)k/r$ and $p \geq s\ln(k)/k$. If the stragglers are selected randomly, then with probability at least $1-\frac{2}{k}$, applying the stochastic block decoding method to a BGC with probability $p$ produces a vector $\vecv$ satisfying
		$$\err(\vecv) \leq \dfrac{16\ln(k)s(1-p)}{p}.$$\end{cor}	

	\subsection{Adversarial Stragglers}\label{sec:adv}

		Suppose now that we have $r$ non-stragglers, but they are selected adversarially by some polynomial-time adversary. When $p = 1$ and $q = 0$, an adversary trying to maximize $\err(\vecv)$ would try to select entire blocks from $\matG$ with the same function assignments. This strategy is efficient for an adversary to compute, even if the adversary views a permuted $\matG$, and leads to a worst-case error of $\err(\vecv) = k-r$ \cite{charles2017approximate}.

		The deterministic block structure of $\matG$ when $p = 1, q = 0$ is what allows an adversary to so easily find a worst-case set of stragglers. In general, the task of selecting $k - r$ stragglers to maximize $\err(\vecv)$ is NP-Hard \cite{charles2017approximate}. When $p = 1$ and $q \ll p$, an adversary could still achieve relatively high error by selecting adversaries from contiguous blocks in the block structure in (\ref{sbc}). This task is equivalent to that of {\it community detection} in the stochastic block model.

		\begin{definition}We say that a community detection algorithm has accuracy $\alpha \in [0,1]$ if at most $(1-\alpha)k$ nodes are misclassified with probability tending to 1 as $k \to \infty$.\end{definition}

		 We say that a community detection algorithm has {\it exact recovery} if it has accuracy $\alpha = 1$ and that it has {\it weak recovery} if it has accuracy $\alpha = \frac{1}{k} + \epsilon$ for $\epsilon > 0$. Weak recovery means that the algorithm performs better than randomly assigning community labels.

		 There has been significant work in understanding when exact and weak recovery are possible. For exact recovery, suppose that we have the stochastic block model in (\ref{sbc}) with $p = a\log(k)/k, q = b\log(k)/k$. Then exact recovery is possible exactly when $\sqrt{a}-\sqrt{b} > \sqrt{k/s}$ \cite{mossel2015reconstruction}. This implies the following theorem.

		\begin{theorem}Suppose we use a stochastic block code $\matG$ with probabilities $p, q$. Then there is no algorithm that can determine the block structure in (\ref{sbc}) with accuracy $\alpha = 1$ as long as $\sqrt{p} - \sqrt{q} < \sqrt{\log(k)/s}.$\end{theorem}

		Thresholds for weak recovery are also known, but not in as much generality. In \cite{abbe2017community}, the authors define the signal-to-noise ratio $\SNR$ of a community detection problem with $p = a/k, q = b/k$ and $m$ communities by
		$$\SNR = \dfrac{(a-b)^2}{m(a+(m-1)b)}.$$

		Suppose that $p = a/k, q = a/k$ and we have 2 communities. Then weak reconstruction is possible exactly when $(a-b)^2 > 2(a+b)$ \cite{abbe2015community}. For $m$ communities, weak recovery is possible when $(a-b)^2 > m(a+(m-1)b).$ However, the converse does not necessarily hold \cite{abbe2017community}. In general, community detection seems to be more difficult for smaller values of $(a-b)^2/m(a+(m-1)b)$. This leads to the following theorem.
		\begin{theorem}An adversary can determine the block structure in (\ref{sbc}) with accuracy better than random iff weak recovery is possible. In particular, for $s = \frac{k}{2}$, this is possible if and only if $k(p-q)^2 > 2(p+q).$\end{theorem}
		
		We would like to note that although there exist regimes for which identifying the community structure in a SBC can be computationally intractable, we have not identified a set of parameters such that worst case straggler detection is NP-hard for SBCs, while average case reconstruction error is small. We however think that the connection to community detection is a very interesting direction that can lead to such results.

	\section{Empirical Results}

		In this section, we compare the empirical error of SBCs for different settings of $p$ and $q$ using the stochastic block decoding method and using the optimal decoding method. We are interested in how the error changes as a function of the number of non-stragglers $r$. Recall that the error of a decoding vector $\vecv$ is given by
		$\err(\vecv) = \|\matG\vecv-\ones_k\|_2^2.$
		We compare this to the uncoded error, which equals the fraction of stragglers.

\begin{figure}[h]
			\centering
			\captionsetup{width=0.5\textwidth}
			\begin{subfigure}{.25\textwidth}
			  \centering
			  \includegraphics[width=\linewidth]{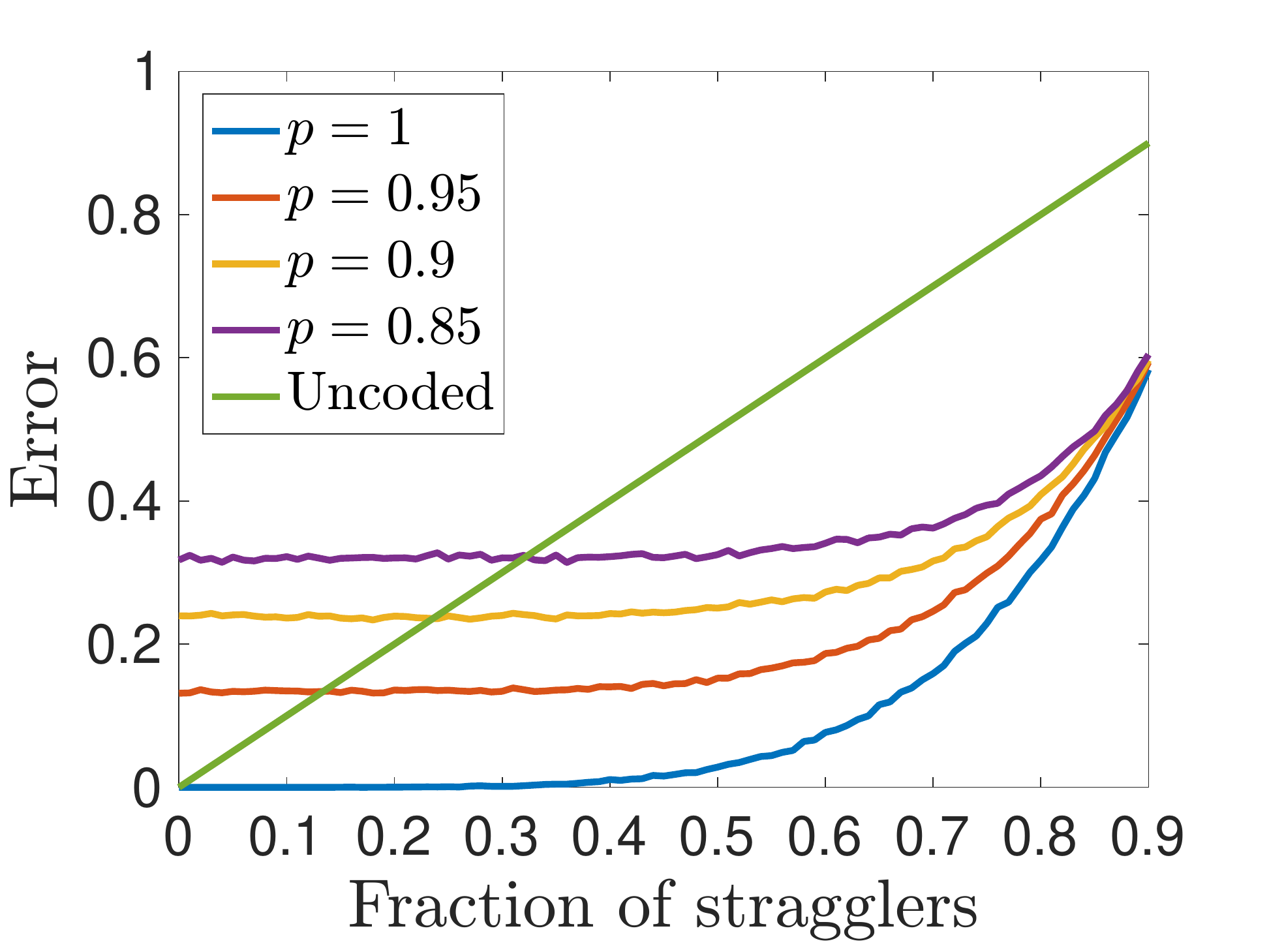}
			  \caption{$s = 5$}
			  \label{fig:err1_s5}
			\end{subfigure}%
			\begin{subfigure}{.25\textwidth}
			  \centering
			  \includegraphics[width=\linewidth]{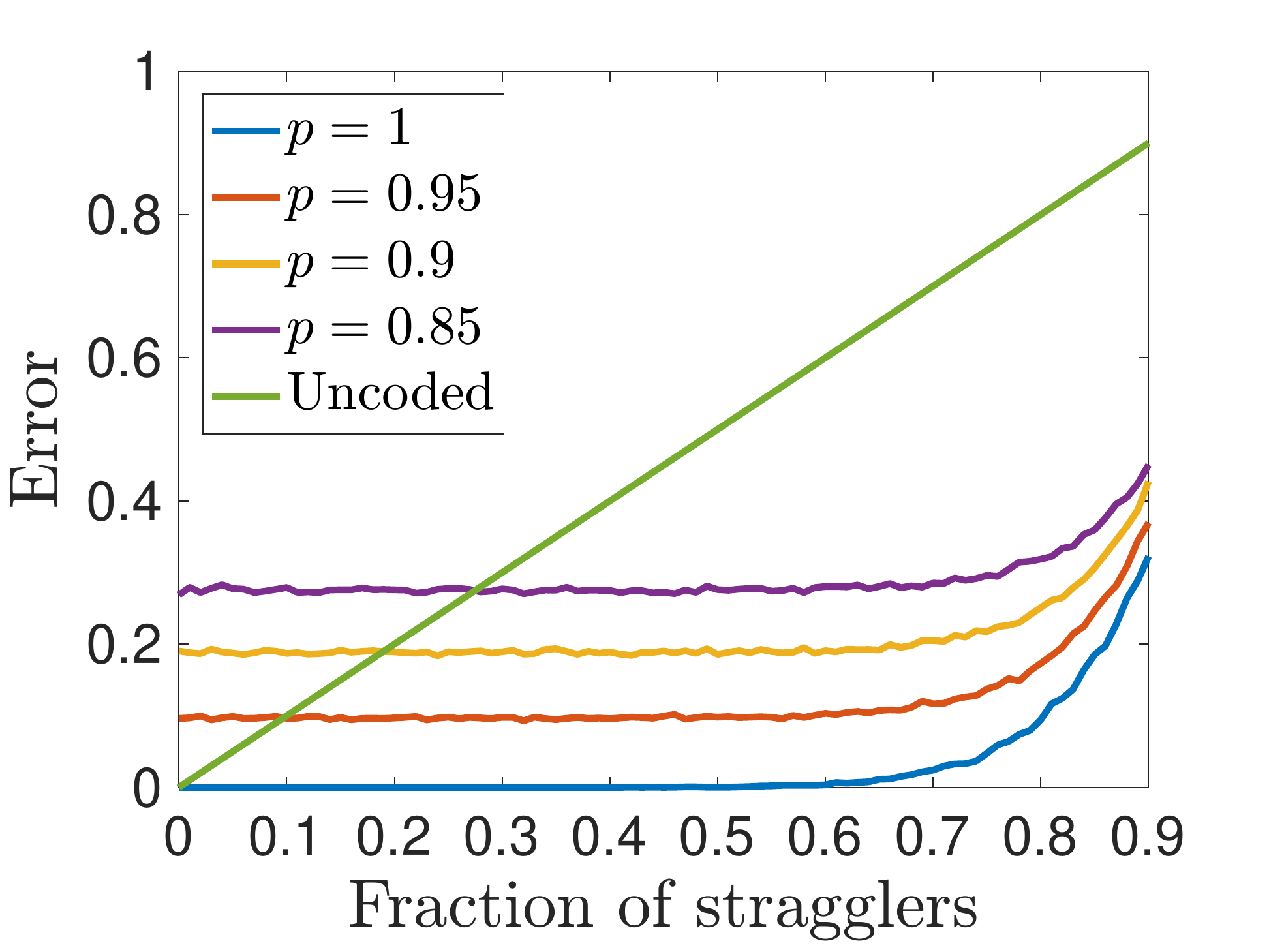}
			  \caption{$s = 10$}
			  \label{fig:err1_s10}
			\end{subfigure}
			\caption{\small A plot of the average error $\err(\vecv)/k$ over 5000 trials. We take $k = 100$, $r = (1-\epsilon)k$ for varying fractions of stragglers $\epsilon$. The figure on the left has $s = 5$ while the figure on the right has $s = 10$.}
			\label{fig:1}
		\end{figure}

		In Figure \ref{fig:1}, we plot $\err(\vecv)/k$ where $\vecv$ comes from stochastic block decoding. We use varying values of $p$ and set $q$ so that the expected number of non-zero entries equals $s$. As our theory suggests, the error increases roughly proportionally to $(1-p)p$ for reasonable $\epsilon$. As $\epsilon$ approaches $1$, $p$ has a smaller effect on the error.

\begin{figure}[h]
			\centering
			\captionsetup{width=0.5\textwidth}
			\begin{subfigure}{.25\textwidth}
			  \centering
			  \includegraphics[width=0.93\linewidth]{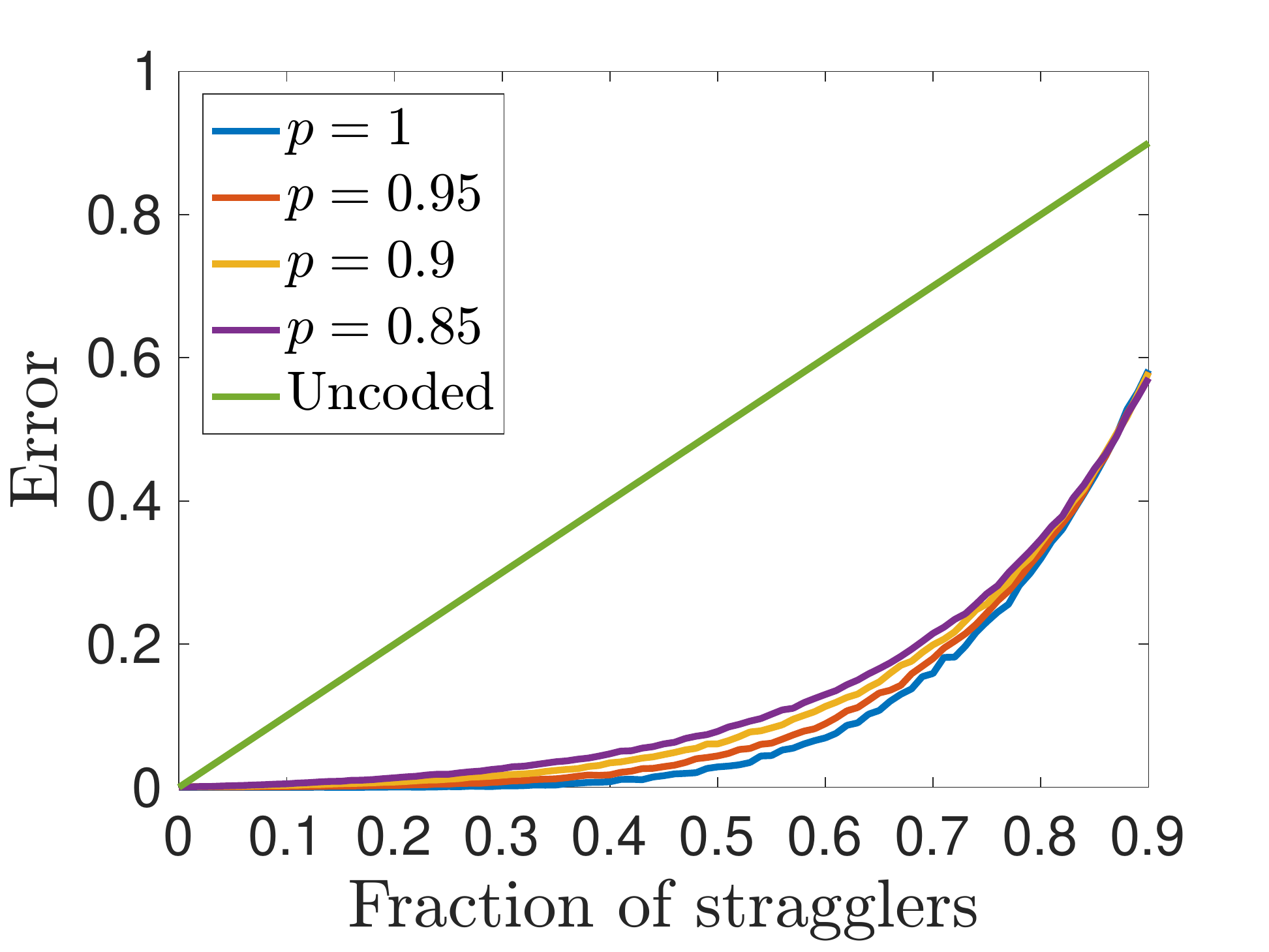}
			  \caption{$s = 5$}
			  \label{fig:err1_s5}
			\end{subfigure}%
			\begin{subfigure}{.25\textwidth}
			  \centering
			  \includegraphics[width=\linewidth]{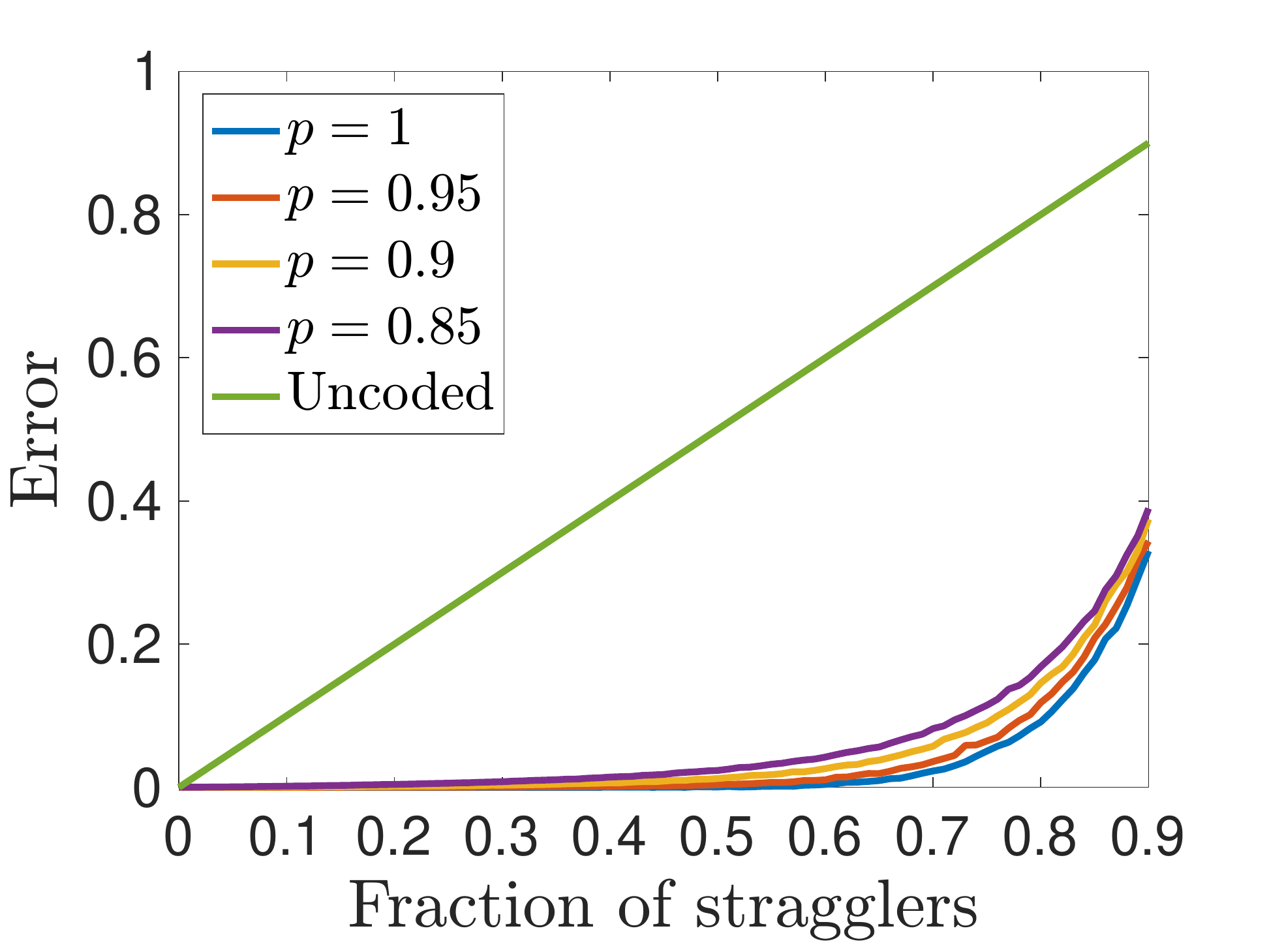}
			  \caption{$s = 10$}
			  \label{fig:err1_s10}
			\end{subfigure}
			\caption{\small A plot of the average error $\err(\vecv_{\text{opt}})/k$ over 5000 trials. We take $k = 100$, $r = (1-\epsilon)k$ for varying fractions of stragglers $\epsilon$. The figure on the left has $s = 5$ while the figure on the right has $s = 10$.}
			\label{fig:2}
		\end{figure}

		In Figure \ref{fig:2}, we plot $\err(\vecv_{\text{opt}})$ for the same values of $p$ and $s$. In this case, the error is less sensitive to $p$ than in stochastic block decoding. Moreover, even when $p = 0.85$, we achieve substantially smaller error than in the uncoded case.

		\vspace{-0.5cm}

\section{Conclusion}

	In this work, we presented a new gradient code, the stochastic block code. The code is efficiently computable and we provided an efficient decoding method for approximate gradient recovery. The code can be tuned according to whether we care more about random or adversarial stragglers and interpolates between previously proposed codes like FRCs and BGCs. We gave theoretical bounds on the error of SBCs and showed empirically that it can achieve close to zero error in the presence of large numbers of random stragglers. We also show that SBCs can be more robust to adversarial stragglers than FRCs.

	Our main results consider a simplified decoding method. In some settings, optimal decoding may be computationally feasible. Understanding optimal decoding error could lead to improved gradient codes. Other open questions involve determining lower bounds on the error of gradient codes and finding optimal codes for a given column sparsity of the function assignment matrix.

\bibliographystyle{IEEEtran}

\bibliography{citations}

\appendices

	\section{Proof of Main Results}

		We wish to prove the results in Section \ref{sec:random_stragglers}. Recall that we have a $k\times k$ function assignment matrix $\matG$ with structure as given in \eqref{sbc}. We only have $r$ non-straggler columns to use in the stochastic block decoding method in Algorithm \ref{alg:sbc_decode}. We let $T$ denote the indices of the non-stragglers. Let $S = \{1,\ldots, k\}$ and let $S_1 = \{1,\ldots, s\}, S_2 = \{s+1,\ldots, 2s\},\ldots, S_{k/s} = \{k-s+1,\ldots, k\}$ and define $T_i := S_i \cap T$. We first wish to bound the probability that each $T_i$ is non-empty. This will allow us to get better bounds on the decoding error, as achieving zero error in the FRC case is only possible when each $T_i \neq \emptyset$.

		\begin{lemma}\label{lem:prob_empty}
		$$\PP\left(\forall i,~T_i \neq \emptyset\right) \geq 1-\dfrac{k}{s}\dfrac{\binom{k-s}{r}}{\binom{k}{r}}.$$\end{lemma}

		\begin{IEEEproof}Fix $i$. Note that $T_i = \emptyset$ iff all $r$ non-stragglers are selected from $S\backslash S_i$, which has size $k-s$. Therefore,
		$$\PP(T_i = \emptyset) = \dfrac{\binom{k-s}{r}}{\binom{k}{r}}.$$
		Taking a union bound over all $T_i$, we arrive at the desired result.\end{IEEEproof}

		This probability is unfortunately unwieldy to use. When $s$ is large enough, however, we can greatly simplify this probability, as in the following lemma.

		\begin{lemma}\label{lem:prob_empty_simpler}
		Suppose that $s \geq 2\ln(k)k/r$. Then 
		$$\PP\left(\forall i,~T_i \neq \emptyset\right) \geq 1-\frac{1}{k}.$$\end{lemma}
		\begin{IEEEproof}By Lemma \ref{lem:prob_empty}, we have
		$$\PP(\exists i,~T_i = \emptyset) \leq \frac{k}{s}\dfrac{\binom{k-s}{r}}{\binom{k}{r}}.$$
		Manipulating, we get
		\begin{align*}
		\dfrac{\binom{k-s}{r}}{\binom{k}{r}} &= \dfrac{(k-s)(k-s-1)\ldots (k-s-r+1)}{k(k-1)\ldots (k-r+1)}\\
		&\leq \left(\dfrac{k-s}{k}\right)^r.\end{align*}
		Therefore,
		$$\PP(\exists i,~T_i = \emptyset) \leq k\left(\dfrac{k-s}{k}\right)^r.$$
		The right-hand side is at most $\frac{1}{k}$ if
		\begin{equation}\label{suff_cond_frc_1}
		\dfrac{s}{k} \geq 1-k^{-2/r}.\end{equation}
		Since $s \geq 2\ln(k)k/r$, we have
		\begin{equation}\label{s_bound}
		\dfrac{s}{k} \geq \dfrac{2\ln(k)}{r}.\end{equation}
		Letting $t = 2\ln(k)/r$, \eqref{s_bound} implies that \eqref{suff_cond_frc_1} holds if
		\begin{align*}
		t \geq 1-e^{-t}.\end{align*}
		Since this occurs for all $t \geq 0$, the desired result is shown.
		\end{IEEEproof}

		In the following, we will let $\mcA$ denote the event that $T_i \neq \emptyset$ for all $i$.

		By Lemma \ref{lem:prob_empty_simpler}, with probability at least $1-\frac{1}{k}$, each $T_i \neq \emptyset$. Therefore, the vector $\vecv$ has an entry of $\frac{1}{\beta}$ in $\frac{k}{s}$ locations, each corresponding to a different block of $\matG$. Let $\vecw_1,\ldots, \vecw_{k/s}$ denote these columns of $\matG$, and let $\vecw := \matG\vecv$. Therefore,
		$$\matG\vecv = \frac{1}{\beta}\sum_{i=1}^{k/s}\vecw_i.$$
		This implies
		\begin{equation}\label{w_eq}
		\err(\vecv) = \|\matG\vecv-\ones_k\|_2^2 = \left\|\frac{1}{\beta}\sum_{i=1}^{k/s}\vecw_i - \ones_k\right\|_2^2.\end{equation}

		We now consider two different regimes of $q$. First, we will analyze the setting where $q = \frac{\gamma}{k}$ for $\gamma < s$. We will show that in this setting, we can analyze the entries of $\matG\vecv$ directly. We first show that for this regime of $q$, the entries of $\matG\vecv$ are not too large.

		\begin{lemma}\label{lem:l_inf}Suppose that $1-p \leq k^{-2}$ and $\gamma < s$. Then
		$$\PP\left(\forall i,~\|\matG\vecv-\ones_k\|_\infty > 7\sqrt{\frac{\ln(k)\gamma}{s}}\bigg|\mcA\right) \leq \frac{2}{k}.$$\end{lemma}

		\begin{IEEEproof}Conditioning on $\mcA$, we have $\matG\vecv = \frac{1}{\beta}\sum_i \vecw_i$ as in \eqref{w_eq}. By the block structure of $\matG$ in \eqref{sbc}, we therefore know that each entry of $\matG\vecv$ is the sum of a $\ber(p)$ random variable and $\frac{k}{s}-1$ $\ber(q)$ random variables, divided by $\beta$. Since $\beta \geq 1$, it suffices to show that with high probability, for each entry, the $\ber(p) = 1$ and that the sum of the corresponding $\ber(q)$ random variables is not too large. Note that the probability that some $\ber(p)$ equals 0 is, by the union bound, at most $k(1-p)$. By assumption on $p$, this is at most $\frac{1}{k}$.

		We now bound the probability of having too many non-zero $\ber(q)$ Let $X$ be the sum of $\frac{k}{s}$ $\ber(q)$ random variables. Note that $\EE[X] = \frac{kq}{s} = \frac{\gamma}{s}$. By the Chernoff bound, $X > 7\sqrt{\ln(k)\gamma/s} > \mu+6\sqrt{\ln(k)\mu}$ with probability at most $\frac{1}{k^2}.$ This implies that the $i$th entry of $\matG\vecv$ is less than $7\sqrt{\ln(k)\gamma/s}$ with the same probability, and so the $i$th entry of $\matG\vecv-\ones_k$ is at most $7\sqrt{\ln(k)\gamma/s}$ with the same probability (conditioned on $\mcA$). Taking a union bound over all $k$ entries, and with the probability that all the $\ber(p) = 1$ as above, we get the desired result.
		\end{IEEEproof}

		We will now prove Theorem \ref{thm:main_thm}.

		\begin{IEEEproof}[Proof of Theorem \ref{thm:main_thm}]Again, conditioning on $\mcA$, we have that $\matG\vecv = \frac{1}{\beta}\sum_i \vecw_i$ as in \eqref{w_eq}. Note that since $q = \frac{\gamma}{k} < \frac{s}{k}$, we know that $p+(\frac{k}{s}-1)q < 2$ and so $\beta = 1$. By the block structure in \eqref{sbc}, this implies that the $i$th entry of $\matG\vecv$ is the sum of a $\ber(p)$ random variable and $\frac{k}{s}-1$ $\ber(q)$ random variables. We wish to bound the number of non-zero entries of $\matG\vecv-\ones_k$.

		Let $X$ denote the $i$th entry of $\matG\vecv-\ones_k$. Then a sufficient condition for $X = 0$ is that the $\ber(p)$ is 1 and all the $\ber(q)$ are 0. We will bound the probability that this happens. For simplicity of notation, we will consider the case where there are $k/s$ $\ber(q)$ random variables. Clearly, the $\ber(p)$ is 0 with probability $1-p$, and the sum of the $\ber(q)$ is larger than 0 with probability $1-(1-q)^{k/s}$. By Bernoulli's inequality \cite{mitrinovic2013classical}, this is at most $\frac{kq}{s} = \frac{\gamma}{s}$. Therefore,
		\begin{equation}\label{eq:nonzero_prob}
		\PP\left((\matG\vecv-\ones_k)_i \neq 0\bigg| \mcA\right) \leq 1-p + \frac{\gamma}{s}.\end{equation}
		Therefore, the expected number of non-zero entries of $\matG\vecv-\ones_k$ is at most $\mu = k(1-p+\frac{\gamma}{s})$. Let $\alpha = (1-p+\frac{\gamma}{s})$. We will consider two cases, depending on $\alpha$.

		{\it Case 1:} $\alpha \geq 3\ln(k)/k$.

		We now wish to bound the number of non-zero entries $N$ in $\matG\vecv-\ones_k$ with high probability. Since the entries are independent and any given entry is non-zero with probability at most $\alpha$, the Chernoff bound implies that for any $\delta \in [0,1]$, $N \geq (1+\delta)k\alpha$ with probability at most $\exp(-\delta^2k\alpha/3)$. Taking $\delta = 1$, this implies that
		\begin{align*}
		\PP\left(N \geq 2k\alpha \bigg|\mcA\right) \leq e^{-k\alpha/3}\leq \frac{1}{k}.\end{align*}
		The second inequality follows from the fact that $\alpha \geq 3\ln(k)/k$. Therefore,
		\begin{equation}\label{eq:case1}
		\PP\left(\|\matG\vecv-\ones_k\|_0 \geq 2k\alpha \bigg| \mcA\right) \leq \frac{1}{k}.\end{equation}

		{\it Case 2:} $\alpha < 3\ln(k)/k$.

		Again, we wish to bound the number of non-zero entries $N$ in $\matG\vecv-\ones_k$ with high probability, again using the Chernoff bound. We now use the fact that for $\delta > 1$, $N \geq (1+\delta)k\alpha$ with probability at most $\exp(-\delta k\alpha/3)$. Let $\delta = 3\ln(k)/k\alpha$. By assumption on $\alpha$, $\delta > 1$. Plugging this $\delta$ in to the Chernoff bound implies
		$$\PP\left( N \geq (1+\delta)k\alpha \bigg| \mcA\right) \leq \frac{1}{k}.$$
		Note that $(1+\delta)k\alpha = k\alpha+3\ln(k) \leq 6\ln(k)$ by assumption on $\alpha$. Therefore,
		\begin{equation}\label{eq:case2}
		\PP\left(\|\matG\vecv-\ones_k\|_0 \geq 2k\alpha \bigg| \mcA\right) \leq \frac{1}{k}.\end{equation}

		Let $M = 14\sqrt{\ln(k)\gamma/s}\max\left\{k\left((1-p)+\frac{\gamma}{s}\right),3\ln(k)\right\}$. Combining Lemma \ref{lem:l_inf} with \eqref{eq:case1} and \eqref{eq:case2}, we get
		$$\PP\left(\|\matG\vecv-\ones_k\|_2^2 \geq M \bigg| \mcA\right) \leq \frac{2}{k}.$$

		Since $\mcA$ holds with probability at least $1-\frac{1}{k}$ a straightforward conditional probability calculation shows
		$$\PP\left(\|\matG\vecv-\ones_k\|_2^2 \geq M\right) \geq 1-\frac{3}{k}.$$\end{IEEEproof}

		Corollary \ref{cor:frc} follows by setting $p = 1$ and noting that when $\gamma \leq 3\ln(k)s/k$, the maximum in Theorem \ref{thm:main_thm} equals 3$\ln(k)$. 

		Next, we consider the setting where $q \gtrapprox s\ln(k)/k$. In order to prove Theorem \ref{thm:main_thm2}, we will use a version of the vector Bernstein inequality to derive high probability bounds on $\err(\vecv)$. We will use a simplified version which can be found in \cite{candes2011probabilistic}.

		\begin{theorem}[Vector Bernstein inequality]\label{thm:vec_bernstein}Let $\{\vecv_i\}_{i=1}^N$ be a finite sequence of independent random vectors. Suppose that $\|\vecv_i-\EE\vecv_i\|_2 \leq B$ a.s. and let $\sigma^2 = \sum_{i=1}^N \EE\|\vecv_i-\EE\vecv_i\|_2^2$. Then for all $t$ satisfying $0 \leq t \leq \sigma^2/B$,
		$$\PP\left(\left\|\sum_{i=1}^N \vecv_i-\EE\vecv_i\right\|_2 > t \right)\leq \exp\left(-\dfrac{t^2}{8\sigma^2}+\frac{1}{4}\right).$$\end{theorem}

		We will now state and prove a more general version of Theorem \ref{thm:main_thm2}. Recall that in stochastic block decoding, we scale by $\beta = p+(\frac{k}{s}-1)q$ if this quantity is at least 2.

		\begin{theorem}\label{thm:ref}Suppose that $s \geq 2\ln(k)k/r$ and $(\frac{k}{s}-1)(1-q)q \geq \ln(k)$. With probability at least $1-\frac{2}{k}$, applying stochastic block decoding to a SBC produces a vector $\vecv$ satisfying
		$$\err(\vecv) \leq \dfrac{16k\ln(k)}{\beta^2}\left((1-p)p + \left(\frac{k}{s}-1\right)(1-q)q\right).$$\end{theorem}

		\begin{IEEEproof}Suppose that event $\mcA$ holds, that is, all $T_i \neq \emptyset$.
		Let the vectors $\vecw_1,\ldots, \vecw_{k/s}$ be the columns corresponding to the non-zero entries of $\vecv$ (provided each $T_i \neq \emptyset$), as in \eqref{w_eq}. Note each $\vecw_i$ has expected value given by a distinct column of the matrix $\matG$ in \eqref{sbc}. In particular, for a given $i$, $\EE\vecw_i$ has $s$ entries of $p$ and $k-s$ entries of $q$. By the block structure in \eqref{sbc}, this implies
		$$\frac{1}{\beta}\sum_{i=1}^{k/s}\EE\vecw_i = \frac{1}{\beta}\left(p + \left(\frac{k}{s}-1\right)q\right)\ones_k = \ones_k.$$
		Therefore,
		\begin{equation}\label{eq:err_comp}
		\err(\vecv) = \left\|\frac{1}{\beta}\sum_{i=1}^{k/s}\vecw_i-\EE\vecw_i\right\|_2^2= \frac{1}{\beta^2}\left\|\sum_{i=1}^{k/s}\vecw_i-\EE\vecw_i\right\|_2^2.\end{equation}
		Note that $\|\vecw_i-\EE\vecw_i\|_2 \leq \sqrt{k}$ but by direct computation,
		$$\sigma^2 = \sum_{i=1}^{k/s}\EE\|\vecw_i-\EE\vecw_i\|_2^2 = k\left((1-p)p+\left(\frac{k}{s}-1\right)(1-q)q\right).$$

		Let $t = 4\sigma\sqrt{\ln(k)}$. Note that if $(\frac{k}{s}-1)(1-q)q \geq \ln(k)$, then $t \leq \sigma^2/\sqrt{k}$, so we can apply Theorem \ref{thm:vec_bernstein}. This then implies that
		\begin{equation}\label{eq:prob_bound}
		\PP\left(\left\|\sum_{i=1}^{k/s}\vecw_i-\EE\vecw_i\right\|_2^2 > t^2 \bigg| \mcA \right) \leq \frac{1}{k}.\end{equation}
		Note that this was conditioned on the event $\mcA$ that $T_i \neq \emptyset$ for all $i$. Combining \eqref{eq:err_comp} and \eqref{eq:prob_bound}, we find that if $\mcA$ holds, then with probability at least $1-\frac{1}{k}$,
		$$\err(\vecv) \leq \dfrac{16k\ln(k)}{\beta^2}\left((1-p)p + \left(\frac{k}{s}-1\right)(1-q)q\right).$$
		Therefore, the probability that $\mcA$ holds and that this bound on $\err(\vecv)$ holds is at least $(1-1/k)^2 \geq 1-2/k$.
		\end{IEEEproof}

		Note that as long as $p \geq q$, $\beta \geq \frac{kq}{s}$. Plugging this in to Theorem \ref{thm:ref}, we derive Theorem \ref{thm:main_thm2}. Setting $p = q$, we arrive at Corollary \ref{cor:bgc}.

\end{document}